\documentclass[conference]{IEEEtran}
\IEEEoverridecommandlockouts
\usepackage{cite}
\usepackage{float} 
\usepackage{graphicx}  
\usepackage{hyperref}

\usepackage{etoolbox}
\usepackage{amsmath,amssymb,amsfonts}
\usepackage{array}
\usepackage{graphicx}
\graphicspath{./image/}
\usepackage{textcomp}
\usepackage{xcolor}
\usepackage{amsmath}
\usepackage{tikz}
\usepackage{nth}
\usepackage{subcaption} 
\usepackage{fancyhdr}
\usepackage{float}
\usepackage{multirow} 
\usepackage[table,xcdraw]{xcolor}
\usepackage{colortbl}
\usepackage{pgf} 

\definecolor{lightgreen}{rgb}{0.85,1.0,0.85}
\definecolor{lightyellow}{rgb}{1.0,1.0,0.6}
\definecolor{lightred}{rgb}{1.0,0.8,0.8}
\usepackage[table]{xcolor}

\definecolor{lightgreen}{RGB}{210,245,210}
\definecolor{lightred}{RGB}{255,220,220}
\usepackage{multirow}
\newcommand{\colcell}[1]{%
  \pgfmathparse{#1*100}%
  \ifdim \pgfmathresult pt < 50pt \cellcolor{lightred}#1%
  \else
    \ifdim \pgfmathresult pt < 85pt \cellcolor{lightyellow}#1%
    \else \cellcolor{lightgreen}#1%
    \fi
  \fi
}

\usepackage{algorithmicx,algpseudocode}
\usepackage{microtype}
\usepackage{algpseudocode}
\usepackage{float}

\usepackage{siunitx,array,multirow}
\usepackage{booktabs}

\usepackage{tikz}
\usetikzlibrary{shapes.geometric, arrows}

\tikzstyle{startstop} = [rectangle, rounded corners, minimum width=3cm, minimum height=1cm,text centered, draw=black, fill=gray!20]
\tikzstyle{process} = [rectangle, minimum width=3.5cm, minimum height=1cm, text centered, draw=black, fill=blue!10]
\tikzstyle{arrow} = [thick,->,>=stealth]

\usepackage{tabularx}
\usepackage{multirow}
\usepackage{threeparttable}
\usepackage{multicol}
\usepackage{xcolor,colortbl}
\usepackage{amsmath}
\usepackage{makecell} 

\usepackage[english]{babel}
\usepackage[utf8]{inputenc}
\usepackage{algorithm}

\usetikzlibrary{shapes,arrows}
\usepackage{verbatim}

\usepackage{url}
\usepackage{amsmath}
\usepackage{textcomp}
\usepackage{siunitx}
\usepackage[utf8]{inputenc}
\usepackage{upgreek}

\makeatletter
 \let\old@ps@headings\ps@headings
 \let\old@ps@IEEEtitlepagestyle\ps@IEEEtitlepagestyle
 \def\confheader#1{%

 \def\ps@IEEEtitlepagestyle{%
 \old@ps@IEEEtitlepagestyle%
 \def\@oddhead{\strut#1\hfill\strut}%
 \def\@evenhead{\strut\hfill#1\hfill\strut}%
 }%
 \ps@headings%
 }
 \makeatother

\begin{document}
\title{Attention-Enhanced Deep Features with Heterogeneous Ensemble Learning for Glaucoma Detection}
\author {\IEEEauthorblockN{Abdullah Al Shafi\IEEEauthorrefmark{1}, Nishat Sadaf Lira\IEEEauthorrefmark{2}, Abrar Hasan\IEEEauthorrefmark{3}, Kazi Saeed Alam\IEEEauthorrefmark{4}, and Swapnil Kundu Argha\IEEEauthorrefmark{5}} 
\IEEEauthorblockA{
Department of CSE, Khulna University of Engineering \& Technology, Bangladesh\IEEEauthorrefmark{1}\IEEEauthorrefmark{4}\IEEEauthorrefmark{5}\\
Department of CSE, Daffodil International University, Bangladesh \IEEEauthorrefmark{2}\\
Department of SWE, Green University of Bangladesh\IEEEauthorrefmark{3}\\
abdullah@iict.kuet.ac.bd\IEEEauthorrefmark{1}, sadaflira.cse@diu.edu.bd\IEEEauthorrefmark{2},
abrar@swe.green.edu.bd\IEEEauthorrefmark{3}, \\ saeed.alam@cse.kuet.ac.bd\IEEEauthorrefmark{4}, swapnilkundu01@gmail.com\IEEEauthorrefmark{5}}}

\maketitle
\begin{abstract}
Glaucoma is a progressive optic neuropathy characterized by irreversible damage to the optic nerve, making timely diagnosis critical to prevent permanent vision loss. Although deep learning has demonstrated promising performance in automated glaucoma detection, existing approaches often overlook feature refinement, suffer from class imbalance, and rely on individual classifiers that limit prediction robustness. To address these challenges, this paper proposes a hybrid glaucoma detection framework that integrates attention-enhanced deep feature extraction with heterogeneous ensemble learning. Specifically, deep representations are extracted using InceptionV3 and subsequently refined by incorporating the Convolutional Block Attention Module (CBAM) to enhance discriminative retinal features. To improve classification robustness, the extracted features are classified using multiple machine learning models together with Single-Level Ensemble (SLE) and Double-Level Ensemble (DLE) strategies, while SMOTE combined with Tomek Links (SMOTE+TL) is employed to alleviate class imbalance. Furthermore, a systematic comparison of handcrafted, deep, and attention-enhanced deep feature representations is conducted. Experimental evaluation on two public retinal fundus datasets demonstrates that deep feature-based methods consistently outperform handcrafted feature-based methods, while the proposed attention-enhanced framework achieves the best overall performance. Furthermore, Grad-CAM visualizations confirm that the proposed model focuses on clinically relevant retinal regions, providing interpretable evidence on the model's prediction process.
\end{abstract}


\begin{IEEEkeywords}
glaucoma diagnosis, attention mechanism, ensemble learning, class imbalance, Grad-CAM.
\end{IEEEkeywords}

\section{Introduction}
Glaucoma is a type of eye condition that puts pressure on the eye, resulting in damage to the optic nerve. As a slowly progressive disease, the symptoms of glaucoma are often unnoticeable. Glaucoma usually causes blurred vision, eye pain, and, in severe cases, loss of sight. As the second-leading cause of irreversible blindness globally, glaucoma affected approximately 80 million people in 2020, a number projected to rise to 111.8 million by 2040 \cite{shan2024global}. However, early diagnosis can help prevent irreversible vision loss and permanent blindness. The traditional method of glaucoma detection is often based on the expertise of professionals, which is a time-consuming process. It includes perimetry assessment, conducting a dilated eye examination, cup-to-disc (CDR) ratio, pachymetry, tonometry, etc. \cite{daud2021review}. Although traditional detection methods provide an established diagnostic baseline, the detection rate can be improved with the help of automated detection, which will reduce the workload of healthcare professionals \cite{cho2024attention}.

Traditionally, the automated glaucoma detection systems involve two initial stages: feature extraction from retinal fundus images and subsequent classification of these features to identify the presence or absence of the disease \cite{daud2021review}. Key diagnostic indicators extracted from fundus images include the morphology of the optic disc (OD) and optic cup (OC), analysis of blood vessels, and calculation of the cup-to-disc ratio (CDR)\cite{Tulasigeri2016}. If any abnormalities are found in these structures, such as an enlarged OC or a CDR exceeding 0.7, they are the indication of glaucoma risk \cite{zhang2010optic}. Furthermore, there are deep convolutional neural network (CNN) architectures that can extract features from fundus images to detect glaucoma.

The main contributions of this work are summarized below:

i) We incorporate the Convolutional Block Attention Module (CBAM) \cite{guo2022attention} into InceptionV3 \cite{ullah2024glaucoma} to enhance feature representation of retinal fundus images. In addition, we compare traditional handcrafted CDR-based features with deep learning-based features to analyze their effectiveness in glaucoma detection.

ii) Two heterogeneous ensemble strategies, namely Single Level Ensemble (SLE) and Double Level Ensemble (DLE), were utilized to improve classification stability and performance while using SMOTE+TL \cite{khandaker2025handling} to handle the class imbalance issue where applicable.


iii) We employ Gradient-weighted Class Activation Mapping (Grad-CAM) \cite{suara2023grad} visualization to provide interpretability and highlight diagnostically relevant regions in fundus images.



\section{Literature Review}
\label{sec:literature}
The traditional approach of glaucoma diagnosis involves the manual screening of retinal fundus images by experts in the field of ophthalmology \cite{alice2023effect}. This is not only a tedious process but also subjective. Additionally, the shortage of ophthalmology experts and well-equipped labs in developing nations emphasizes the importance of automated tools for the diagnosis of eye diseases \cite{shan2024global}.


According to Daud et al.\cite{daud2021review}, the first attempt at automated glaucoma detection was done using traditional machine learning (ML) techniques, which were primarily based on hand-crafted features. \cite{Tulasigeri2016} proposed a sophisticated thresholding technique for glaucoma diagnosis using Otsu's technique. 



Joshua et al. \cite{Joshua2019} proposed a method for automated diagnosis of glaucoma by accurately measuring CDR using U-net-CNN architecture. Their proposed model performed better than the existing model in the dice-score. Latif et al. \cite{Latif2022-xs} has proposed ODGNet which localized OD followed by employing a transfer learning based pre-trained model. It was tested on ORIGA dataset and found an accuracy of  95.75\%.


In their work, Alice et al. \cite{alice2023effect} proposed an automated system to detect glaucoma from retinal fundus images which involves a Random Forest (RF) classifier combined with various image feature descriptors to classify images. Again, Zhang et al. \cite{zhang2010optic} have proposed an improved method for localizing the OD ROI in retinal fundus images by removal of bright fringes.



Furthermore, Aljohani et al. \cite{aljohani2024hybrid} presented a hybrid approach for glaucoma detection using a combined SVM-ANN classifier for diagnosis, which achieved high accuracy and robustness compared to existing techniques.


Recently, Cho et al. \cite{cho2024attention} proposed a model based on attention mechanism to detect the eye disease using retinal fundus image. They have found that by feeding  preprocessed image into a multi-input CNN with attention mechanisms, the model achieved superior performance in glaucoma classification compared to existing research models.

\section{Dataset Description}
\label{dataset}
Two benchmark datasets, sourced from Kaggle, were utilized for the experiment.

\begin{figure}[htbp]
\centering
     \begin{subfigure}[b]{0.22\textwidth}
         \centering
         \includegraphics[width=\textwidth]{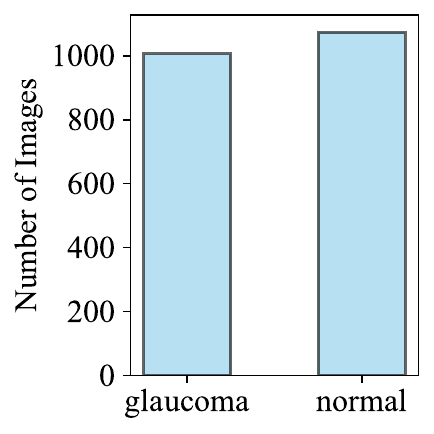}
         \caption{EDC dataset.}
         \label{edc_class_distribution}
     \end{subfigure}
     \begin{subfigure}[b]{0.22\textwidth}
         \centering
         \includegraphics[width=\textwidth]{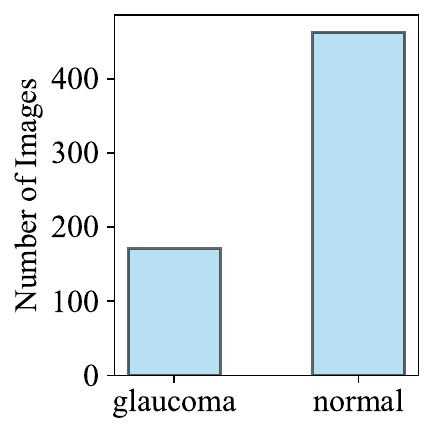}
         \caption{BEH dataset.}
         \label{beh_class_distribution}
     \end{subfigure}
        \caption{Class distribution of the two datasets used in this work.}
\label{edc+beh}
\end{figure}

(1) \textbf{EDC dataset: }This dataset\footnote{https://www.kaggle.com/datasets/gunavenkatdoddi/eye-diseases-classification/data} originally consists of four classes. For our work, we have taken only two classes, namely glaucoma and normal, which contain around 2000 fundus images as shown in Fig. \ref{edc_class_distribution}.


(2) \textbf{BEH dataset: }BEH\footnote{https://www.kaggle.com/datasets/kabirjaan/beh-glaucoma} comprises 634 eye images under two categories (glaucoma and normal). From Fig. \ref{beh_class_distribution}, the dataset is highly imbalanced.

\section{Proposed Glaucoma Detection Method}
\label{methodology}
\begin{figure}[ht]
\centering
\centerline{\includegraphics[width=0.48\textwidth]{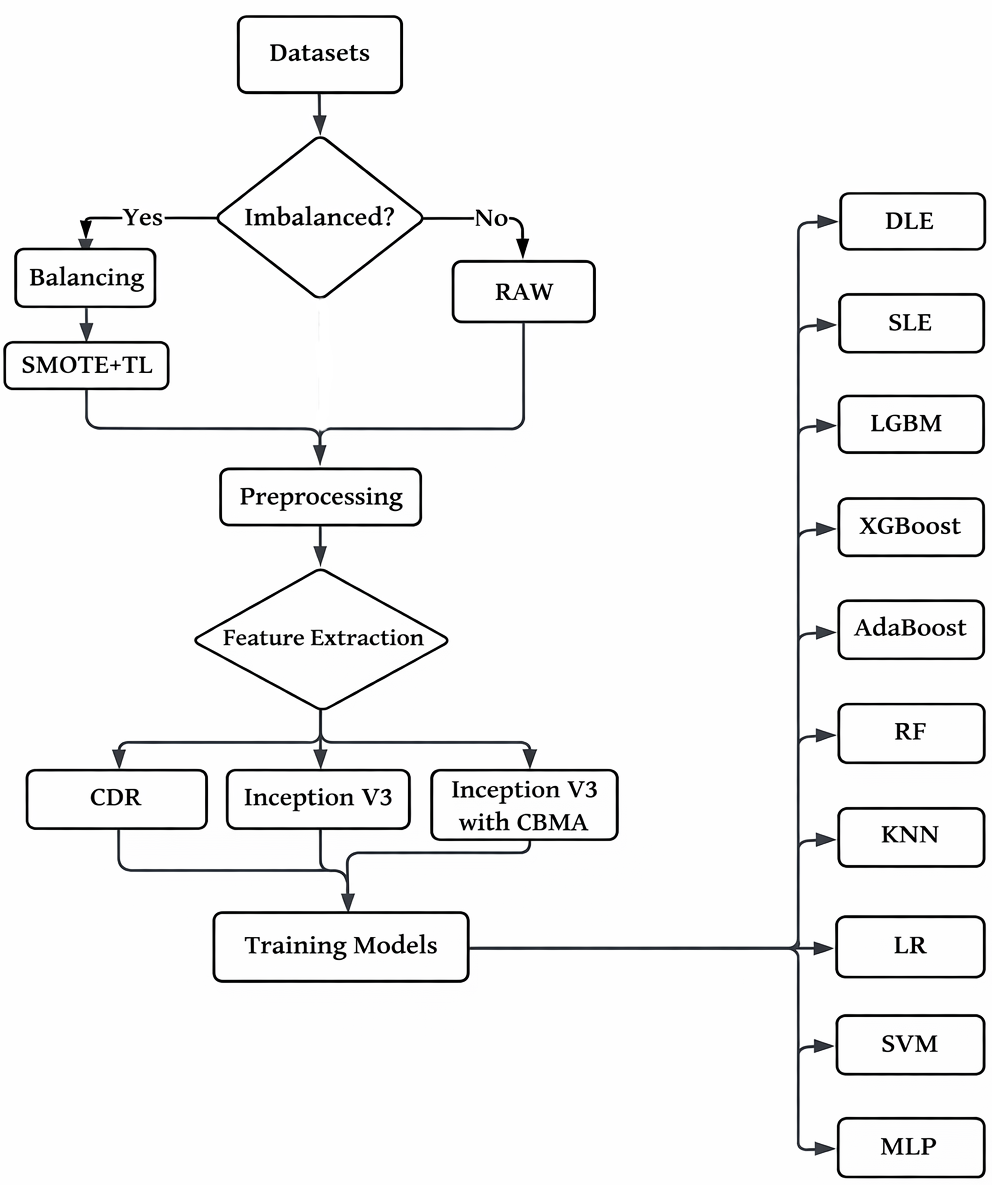}}
\caption{Overall workflow of our proposed glaucoma detection method from the fundus image.}
\label{fig:framework}
\end{figure}

As shown in Fig. \ref{fig:framework}, our proposed architecture consists of several preprocessing steps, hand-crafted and automated feature extraction, and finally classification using traditional ML classifiers along with advanced ensemble strategies.

\begin{figure*}[htbp!]
    \centering
    \begin{subfigure}{0.56\textwidth} 
        \centering
        \includegraphics[width=\textwidth]{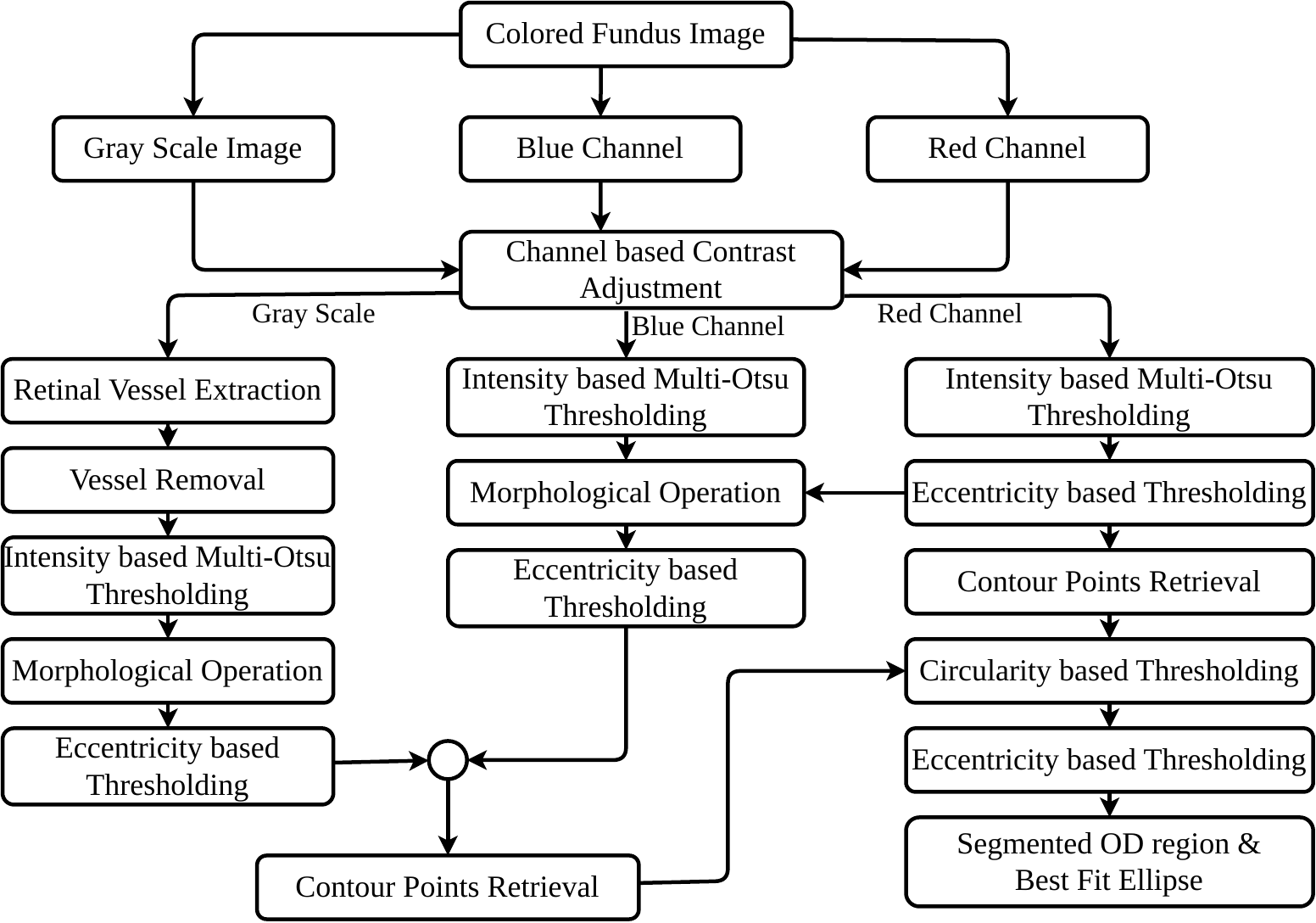}
        \caption{Optic disc segmentation.}
        \label{fig:cdr_od}
    \end{subfigure}%
    \hspace{1em}
    \begin{subfigure}[b]{0.35\textwidth}
        \centering
        \includegraphics[width=\textwidth]{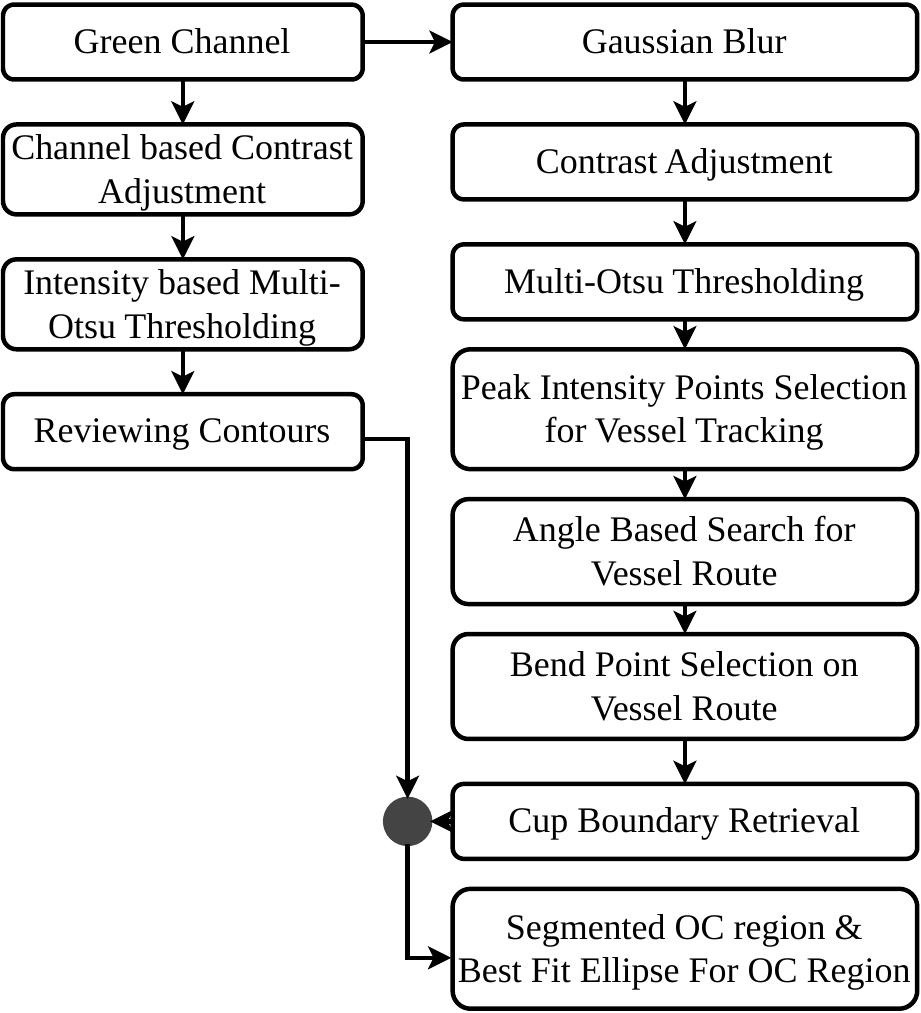}
        \caption{Optic cup segmentation.}
        \label{fig:cdr_oc}
    \end{subfigure}
    \caption{Image processing techniques for CDR calculation.}
    \label{fig:CDR}
\end{figure*}

\begin{figure}[ht]
\centering
\centerline{\includegraphics[width=0.48\textwidth]{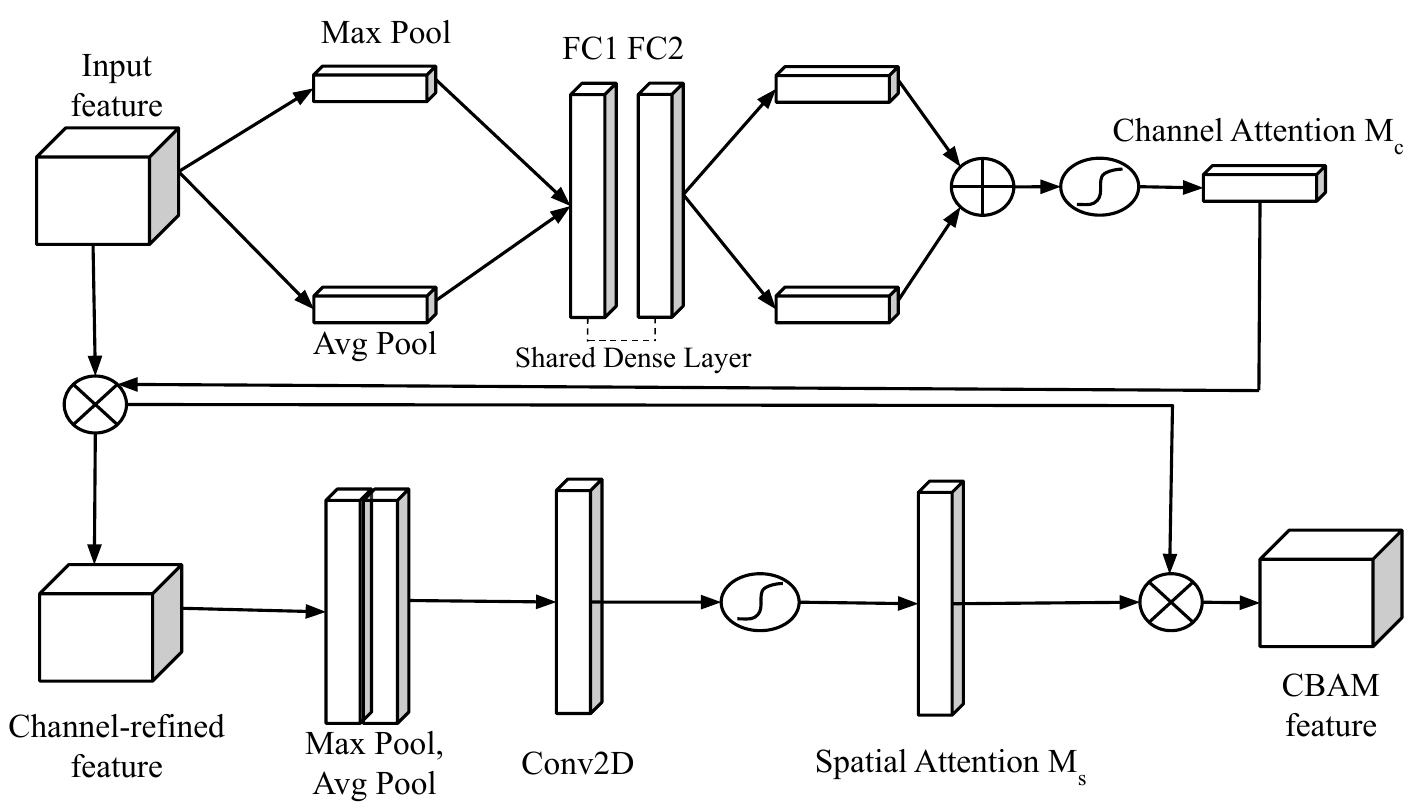}}
\caption{Demonstrating how Convolutional Block Attention Module (CBAM) improves the input features. Here, the input feature refers to the features extracted from InceptionV3.}
\label{fig:cbam}
\end{figure}

\begin{figure}[ht]
\centering
\centerline{\includegraphics[width=.326\textwidth]{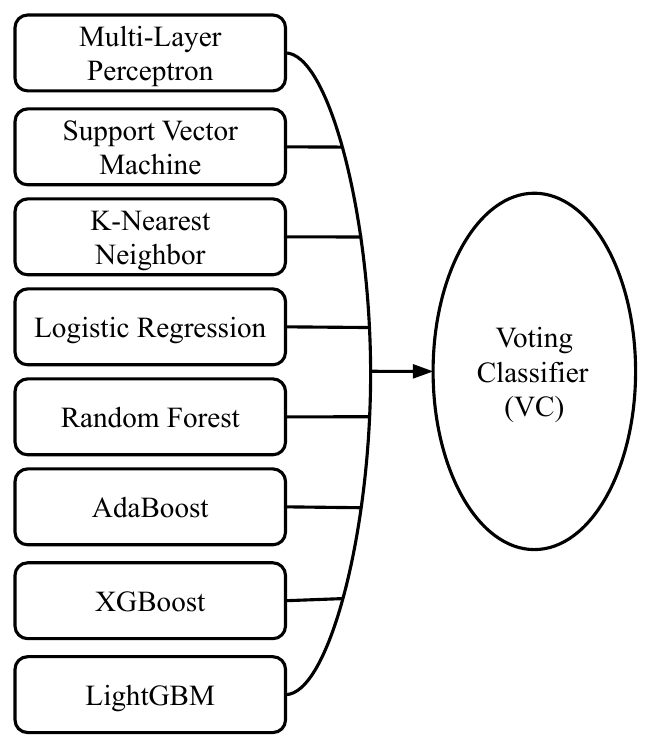}}
\caption{Single Level Ensemble (SLE).}
\label{fig:sle}
\end{figure}

\begin{figure}[ht]
\centering
\centerline{\includegraphics[width=.40\textwidth]{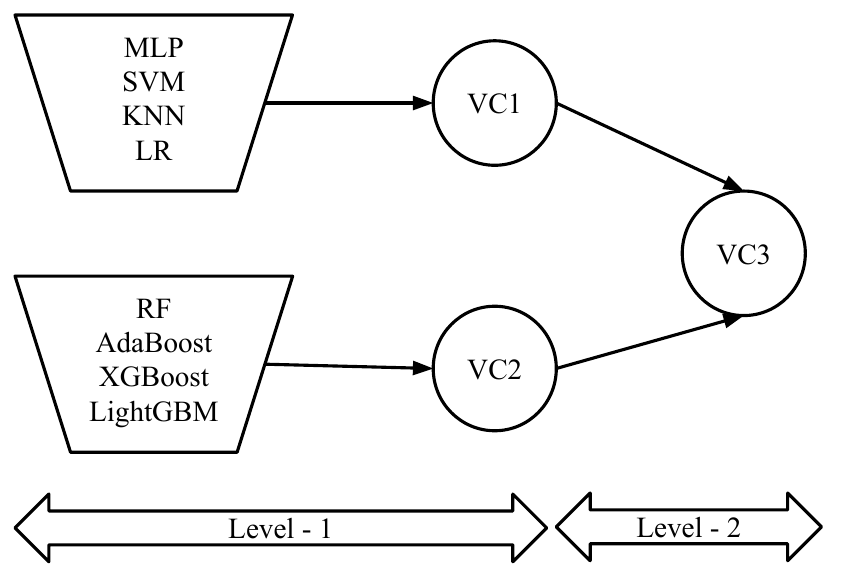}}
\caption{Double Level Ensemble (DLE).}
\label{fig:dle}
\end{figure}


\subsection{Preprocessing}
\textbf{Balancing:} The BEH dataset contains substantially fewer glaucoma images than normal images. This imbalance can cause classifiers to favor the majority class, resulting in poor glaucoma sensitivity. Therefore, SMOTE+TL \cite{khandaker2025handling} was applied to the training set to increase minority-class representation and improve class separability. It should be noted that the validation and test sets have not been subjected to this balancing technique and retain their original class distributions to ensure an unbiased evaluation of model performance.


\textbf{Resizing: }Each of the input fundus images was resized to a dimension of 299x299 pixels, adhering to the requirement of the input size specification of InceptionV3 architecture \cite{ullah2024glaucoma}.


\textbf{Histogram equalization: }We apply histogram equalization to improve the contrast of fundus images.





\textbf{Normalization: }Later, the pixel intensities of the image are rescaled from [0, 255] to [-1, 1]. 




\subsection{Feature Extraction}
\textbf{CDR Calculation: }Initially, the CDR was manually extracted from the fundus image using several image processing techniques. As shown in Fig. \ref{fig:cdr_od}, the image was first separated into color channels. After that, image enhancement was applied, followed by vessel enhancement using the multiscale Frangi filter, and vessel removal using the Telea inpainting method to reduce noise and facilitate OD detection. The OD boundary was then estimated by fitting a best-fit ellipse using the red-channel intensity to obtain its major radius.

After locating the OD, the OC was detected using multi-Otsu thresholding and the bending points of retinal vessels. As illustrated in Fig. \ref{fig:cdr_oc}, intensity-based processing was performed on the green channel, where vessels are more prominent. Low-intensity vessel pixels were detected using a circular window, and their bending points near the OC boundary were used to form a convex hull representing the OC. Finally, the vertical CDR was computed as the ratio of the major radii of the OC and OD.



\begin{table*}[htbp]
\centering
\caption{Performance evaluation of various classification approaches on EDC dataset.}
\resizebox{\textwidth}{!}{
\begin{tabular}{|c|c|c|c|c|c|c|c|c|c|c|c|c|}
\hline
\textbf{Metric} & \textbf{Feature Extraction} & \textbf{MLP} & \textbf{SVM} & \textbf{KNN} & \textbf{LR} & \textbf{RF} & \textbf{Ada} & \textbf{XGB} & \textbf{LGBM} & \textbf{SLE} & \textbf{DLE} \\
\hline
\multirow{3}{*}{Accuracy} & CDR & 0.57 & 0.63 & 0.54 & 0.63 & 0.54 & 0.59 & 0.58 & 0.63 & 0.57 &  0.61 \\
 & InceptionV3 & 0.87 & 0.87 & 0.86 & 0.86 & 0.87 & 0.87 & 0.87 & 0.88 & \textbf{0.89} & 0.88\\
 & \textbf{InceptionV3+CBAM} & 0.87 & 0.88 & 0.86 & 0.86 & 0.87 & 0.87 & \textbf{0.89} & \textbf{0.89} & \textbf{0.90} & \textbf{0.90}\\
\hline
\multirow{3}{*}{Precision} & CDR & 0.48 & 0.31 & 0.49 & 0.42 & 0.48 & 0.50 & 0.48 & 0.31 & 0.50 & 0.58 \\
 & InceptionV3 & 0.87 & 0.87 & 0.86 & 0.86 & 0.87 & 0.87 & 0.87 & 0.88 & \textbf{0.89} & 0.88\\
 & \textbf{InceptionV3+CBAM} & 0.87 & 0.88 & 0.87 & 0.86 & 0.88 & 0.87 & \textbf{0.90} & \textbf{0.89} & \textbf{0.91} & \textbf{0.91}\\
\hline
\multirow{3}{*}{Recall} & CDR & 0.49 & 0.50 & 0.49 & 0.40 & 0.48 & 0.50 & 0.49 & 0.50 & 0.57 & 0.61\\
 & InceptionV3 & 0.87 & 0.87 & 0.86 & 0.86 & 0.87 & 0.87 & 0.87 & 0.88 & \textbf{0.89} & 0.88\\
 & \textbf{InceptionV3+CBAM} & 0.86 & 0.87 & 0.86 & 0.86 & 0.87 & 0.87 & \textbf{0.89} & \textbf{0.88} & \textbf{0.90} & \textbf{0.90}\\
\hline
\multirow{3}{*}{F1-Score} & CDR & 0.52 & 0.39 & 0.49 & 0.40 & 0.47 & 0.46 & 0.46 & 0.39 & 0.52 & 0.59\\
 & InceptionV3 & 0.87 & 0.87 & 0.86 & 0.86 & 0.87 & 0.87 & 0.87 & 0.88 & \textbf{0.89} & 0.88\\
 & \textbf{InceptionV3+CBAM} & 0.87 & 0.88 & 0.86 & 0.86 & 0.87 & 0.87 & \textbf{0.89} & \textbf{0.89} & \textbf{0.90} & \textbf{0.90}\\
\hline
\end{tabular}
}
\label{result_edc}
\end{table*}

\textbf{Deep Learning Method: }The deep learning method utilized in this work for the extraction of features is InceptionV3. Fundus images contain retinal structures at different spatial scales \cite{daud2021review}. The model includes inception modules, which work in parallel with different sizes of convolution filters, thus managing to capture features on different scales \cite{ullah2024glaucoma}.  




\textbf{Deep Learning with Attention Module: }Glaucoma-related structural changes are primarily located around OD and OC. Therefore, merely extracting deep features may not sufficiently emphasize diagnostically relevant regions. To improve the representational capability of the convolutional feature maps of InceptionV3, we have added CBAM as shown in Fig. \ref{fig:cbam}.
CBAM is a light-weight yet efficient attention module that sequentially applies channel attention (CA) and spatial attention (SA) to emphasize informative feature regions while suppressing less useful ones. Taking an intermediate feature map $F$, the channel-refined feature $F'$ and the final attention-refined feature $F''$ are computed as,
\begin{equation}
F' = M_c(F) \otimes F
\end{equation}
\begin{equation}
F'' = M_s(F') \otimes F'
\end{equation}
where $\otimes$ denotes element-wise multiplication, and $M_c$ and $M_s$ represent the channel and spatial attention maps, respectively.


For the construction of the proposed feature extractor, CBAM modules were incorporated into the InceptionV3 network along multiple hierarchical levels. In particular, a CBAM block was added after each of the mixed layers (\texttt{mixed0} through \texttt{mixed9}) in the InceptionV3 network. Additionally, a Global Average Pooling (GAP) layer and a fully connected layer with 1024 neurons using the ReLU activation function were used to produce the final feature representation.


\subsection{Classification Process}
\subsubsection{Traditional ML Models}
We have explored the performance of several conventional ML methods including Multi-Layer Perceptron (MLP), Support Vector Machine (SVM), K-Nearest Neighbor (KNN), Logistic Regression (LR), Random Forest (RF), AdaBoost (Ada), XGBoost (XGB), LightGBM (LGBM) with all extracted features mentioned in the previous subsection.

\subsubsection{Advanced Ensemble Techniques}
Individual classifiers may capture different decision boundaries from the same feature representation. Therefore, combining heterogeneous classifiers can reduce dependence on a single decision function and improve prediction stability. Thus, two heterogeneous ensembles of the above ML techniques have been used to improve accuracy alongside other performance measures.

\textbf{SLE: }A voting classifier was developed using SLE that incorporates all the ML methods used in our classification scheme, which is shown in Fig. \ref{fig:sle}.

\textbf{DLE: }We divided the various classifiers into two groups for the DLE scenario, and two voting classifiers were produced. Fig. \ref{fig:dle} shows the DLE construction method, which comprises two levels (Level - 1, and Level - 2). In Level-1, VC1 and VC2 have been constructed. The VC1 consists of standalone ML models, namely MLP, SVM, KNN, and LR. The VC2 consists of homogeneous ensemble models, namely RF, AdaBoost, XGBoost, LightGBM. And finally in Level - 2, VC1 and VC2 were ensembled to form VC3.

\subsection{Visual Explanation using Grad-CAM}
For interpretation of the model outcome, We used Grad-CAM \cite{suara2023grad} to focus fundus image locations that have the strongest effect on model prediction. Grad-CAM uses gradients from the final convolutional layer to weight feature maps and generate a heatmap indicating the most influential regions for model’s prediction.




\section{Experimental Analysis and Results}
\label{experiment}
\subsection{Experimental Setup and Evaluation Metrics}
The work presented in this paper was done in Python using the CUDA Toolkit for the acceleration of computations on a GPU. Each dataset was divided into three divisions: 80\% training, 10\% validation, and 10\% testing. Evaluation is carried out using macro precision, recall, F1-Score along with accuracy and inference time \cite{al2025skin}. 




\begin{table*}[!htbp]
\centering
\caption{Class-wise performance comparison of various classification approaches on InceptionV3+CBAM features of the BEH dataset. Accuracy is overall, while Precision, Recall and F1-Score are reported per class.}
\resizebox{\textwidth}{!}{%
\begin{tabular}{|c|c|c|c|c|c|c|c|c|c|c|c|c|}
\hline
\textbf{Metric} & \textbf{Balancing} & \textbf{Class} & \textbf{MLP} & \textbf{SVM} & \textbf{KNN} & \textbf{LR} & \textbf{RF} & \textbf{Ada} & \textbf{XGB} & \textbf{LGBM} & \textbf{SLE} & \textbf{DLE} \\
\hline

\multirow{2}{*}{Accuracy} 
 & Imbalanced & Overall & 0.81 & 0.78 & 0.77 & 0.81 & 0.77 & 0.78 & 0.78 & 0.79 & 0.81 & 0.78 \\
 \cline{2-13}
 & SMOTE+TL  & Overall & 0.86 & \textbf{0.89} & 0.74 & 0.86 & 0.89 & 0.87 & 0.88 & \textbf{0.90} & \textbf{0.90} & \textbf{0.89} \\
\hline

\multirow{4}{*}{Precision} 
 & \multirow{2}{*}{Imbalanced} & Glaucoma & 0.71 & 1.00 & 0.56 & 0.65 & 0.67 & 0.64 & 0.61 & 0.63 & 0.73 & 0.62 \\
 &  & Normal  & 0.82 & 0.78 & 0.80 & 0.86 & 0.78 & 0.80 & 0.81 & 0.82 & 0.82 & 0.81 \\
  \cline{2-13}
 & \multirow{2}{*}{SMOTE+TL} & Glaucoma & 0.84 & 0.81 & 0.66 & 0.81 & 0.91 & 0.89 & 0.90 & 0.92 & 0.88 & 0.89 \\
 &  & Normal & 0.89 & 0.86 & 0.96 & 0.92 & 0.88 & 0.86 & 0.87 & 0.89 & 0.91 & 0.89 \\
\hline

\multirow{4}{*}{Recall} 
 & \multirow{2}{*}{Imbalanced} & Glaucoma 
 & 0.39 
 & 0.13 
 & 0.32 
 & 0.55 
 & 0.19 
 & 0.29 
 & 0.35 
 & 0.39 
 & 0.35 
 & 0.32 \\

 &  & Normal & 0.95 & 1.00 & 0.91 & 0.90 & 0.97 & 0.95 & 0.92 & 0.92 & 0.96 & 0.94 \\
  \cline{2-13}

 & \multirow{2}{*}{SMOTE+TL} & Glaucoma 
 & 0.89 
 & 0.87 
 & 0.88 
 & \textbf{0.90} 
 & 0.87 
 & 0.85 
 & 0.86 
 & 0.88 
 & \textbf{0.91} 
 & \textbf{0.89} \\

 &  & Normal & 0.83 & 0.81 & 0.51 & 0.78 & 0.91 & 0.89 & 0.90 & 0.92 & 0.88 & 0.89 \\
\hline

\multirow{4}{*}{F1-Score} 
 & \multirow{2}{*}{Imbalanced} & Glaucoma 
 & 0.50 
 & 0.23 
 & 0.41 
 & 0.60 
 & 0.30 
 & 0.40 
 & 0.45 
 & 0.48 
 & 0.48 
 & 0.43 \\

 &  & Normal   & 0.88 & 0.87 & 0.85 & 0.88 & 0.87 & 0.87 & 0.86 & 0.87 & 0.88 & 0.87 \\
  \cline{2-13}

 & \multirow{2}{*}{SMOTE+TL} & Glaucoma 
 & 0.86 
 & 0.84 
 & 0.79 
 & 0.87 
 & \textbf{0.89} 
 & 0.87 
 & 0.88 
 & \textbf{0.90} 
 & \textbf{0.90} 
 & \textbf{0.89} \\

 &  & Normal & 0.86 & 0.83 & 0.66 & 0.85 & 0.89 & 0.87 & 0.88 & 0.91 & 0.90 & 0.89 \\
\hline

\end{tabular}%
}
\label{tab:classwise_compact}
\end{table*}



\begin{figure}[ht]
\centering
\centerline{\includegraphics[scale=.62]{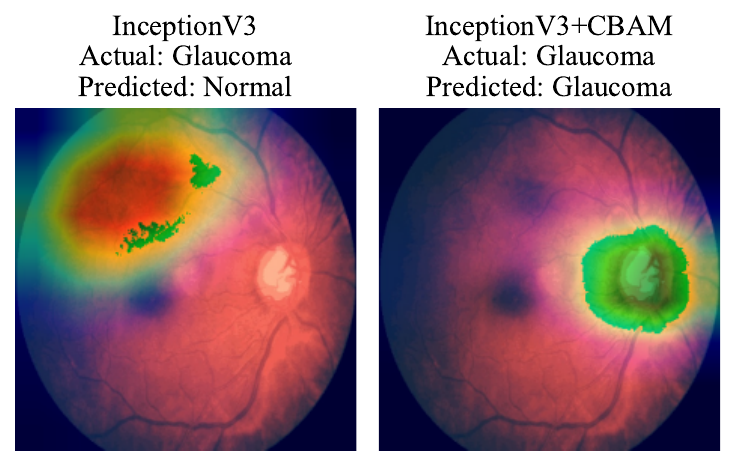}}
\caption{Grad-CAM attention maps illustrating the focus regions of InceptionV3 (misclassified) and InceptionV3+CBAM (correctly classified) feature extractor for glaucoma detection.}
\label{fig:gradcam}
\end{figure}

\subsection{Result Analysis on EDC Dataset}
Table \ref{result_edc} displays the assessment results of various classifiers using three different feature extraction methods to analyze the EDC dataset. The results show that deep feature representations provide superior performance compared to traditional CDR-based features. The CDR feature-trained models exhibited low accuracy along with weak precision-recall performance, which demonstrated the limited identification ability of traditional CDR measurement techniques. InceptionV3 deep features provide substantial performance improvements to all classifiers which achieve accuracy rates between 0.86 and 0.88 and F1-Scores between 0.87 and 0.89. The CBAM attention mechanism improves feature representation which results in the strongest overall performance results. The ensemble approaches show better performance than other methods as the SLE and DLE models achieve their maximum accuracy of 0.90 through InceptionV3+CBAM features while their precision, recall and F1-Score also improved. In case of individual classifiers, gradient boosting methods like XGBoost and LightGBM demonstrate strong performance when they use attention-based features. The results demonstrate that attention-based deep feature extraction together with ensemble learning produces better identification results for glaucoma classification than standard CDR assessment methods.

\begin{figure}[ht]
\centering
\centerline{\includegraphics[width=.40\textwidth]{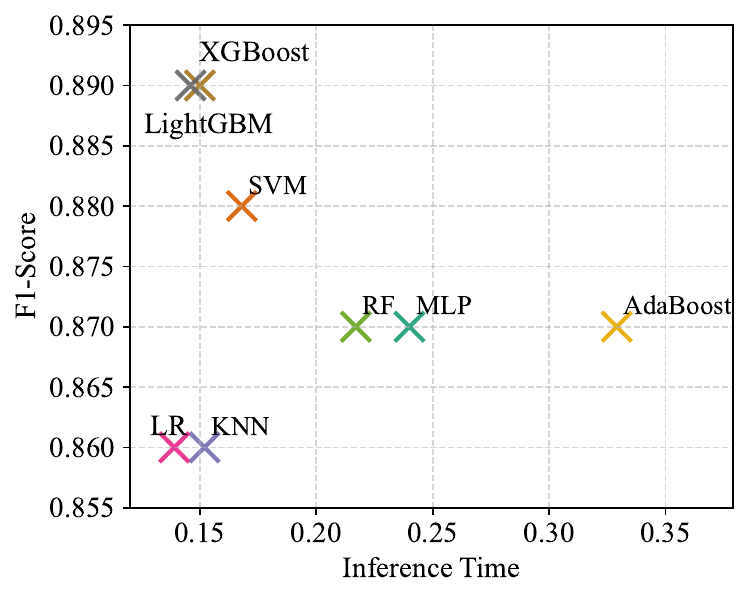}}
\caption{Classifier performance vs total inference time with InceptionV3+CBAM.}
\label{fig:f1vsinf}
\end{figure}

Furthermore, Grad-CAM visualizations in Fig. \ref{fig:gradcam} provide insight into the feature extraction of InceptionV3 and InceptionV3+CBAM for glaucoma detection. In the left panel, corresponding to the InceptionV3, the network incorrectly classifies the image as normal despite the presence of glaucoma. The Grad-CAM heatmap reveals that the model’s attention is widely distributed across the retinal background and peripheral regions rather than being concentrated on the OD area, which is the most clinically relevant region for glaucoma assessment. This diffuse and unfocused activation suggests that InceptionV3 fails to effectively capture discriminative structural features such as the CDR, which are essential indicators of glaucomatous damage. In contrast, the right panel shows the Grad-CAM visualization for the InceptionV3+CBAM model, which correctly predicts the image as glaucoma. Here, the heatmap is more localized and strongly concentrated around the OD and surrounding cup region, indicating that the attention mechanism enables the model to emphasize diagnostically meaningful retinal structures. This focused activation provides qualitative evidence that the CBAM module enhances both channel and spatial feature representation, guiding the network to prioritize informative regions while suppressing irrelevant background information. Consequently, the improved attention distribution contributes to more accurate classification and better alignment with clinically relevant features, highlighting the effectiveness of the CBAM-enhanced architecture for glaucoma detection.

Additionally, Fig. \ref{fig:f1vsinf} shows how different individual classifiers experience a trade-off between their F1 score and total inference time. XGBoost achieves the highest F1 while maintaining relatively low inference time, which demonstrates a favorable trade-off between predictive performance and computational efficiency. LightGBM follows closely with slightly lower performance but comparable efficiency. The combination of AdaBoost's highest inference time and its marginal F1 improvement demonstrates that this method operates less effectively than other individual techniques. The traditional classifiers SVM, RF, MLP, LR, and KNN demonstrate lower F1-Scores in comparison to their competitors, although some classifiers achieve moderate inference times. KNN and LR produce the quickest inference times, but they show the lowest accuracy for their prediction results.

\subsection{Result Analysis on BEH Dataset}
Table \ref{tab:classwise_compact} presents a detailed class-wise evaluation of various classifiers with InceptionV3+CBAM features for the BEH dataset, showing how class imbalance affects the results and how SMOTE+TL augmentation affects the outcomes. Under imbalanced conditions, glaucoma detection is adversely affected by the strong bias toward the majority (normal) class. SMOTE+TL brings major improvements to glaucoma recognition, which enables the models to achieve better performance by increasing the representation of the minority (glaucoma) class. Normal class performance remains relatively stable across both settings, which demonstrates that balancing mainly benefits the minority (glaucoma) class while maintaining majority class accuracy. In short, SMOTE+TL increases overall accuracy, but the main discovery shows that class balancing improves glaucoma sensitivity, thereby improving the clinical reliability of the predictions.

\section{Conclusions}
\label{conclusion}
This paper presented a hybrid glaucoma detection framework that integrates attention-enhanced deep feature extraction with heterogeneous ensemble learning for retinal fundus image analysis. By systematically comparing handcrafted CDR features, InceptionV3-based deep features, and CBAM-enhanced deep features, the study demonstrated the superiority of attention-enhanced deep representations for glaucoma detection. Furthermore, the heterogeneous SLE and DLE ensemble strategies improved classification robustness, while SMOTE+TL effectively mitigated class imbalance, leading to enhanced glaucoma sensitivity on the imbalanced BEH dataset. Experimental results demonstrated that the proposed framework outperformed conventional handcrafted and deep feature-based approaches. In addition, Grad-CAM visualizations provided qualitative evidence that the proposed model focuses on clinically relevant retinal regions, supporting the reliability of its predictions. Thus, the proposed system could assist ophthalmologists by prioritizing suspicious cases for further examination, particularly in settings where access to specialized ophthalmic expertise is limited. However, the experiments were conducted on relatively small public datasets, which may limit the generalizability of the model. Future work will focus on validation using larger datasets and improving computational efficiency.

\section*{Acknowledgment}
This work is supported by CRITS, Green University of Bangladesh.

\bibliographystyle{./IEEEtran}
\bibliography{./IEEEexample}

\end{document}